\pdfoutput=1

\documentclass[11pt]{article}

\usepackage[final]{acl}
\usepackage{amssymb}
\usepackage{graphicx}
\usepackage{comment}
\usepackage{booktabs}
\usepackage{tabularx}
\usepackage{multirow}
\usepackage[table, dvipsnames]{xcolor}
\usepackage{xcolor,soul}
\usepackage{subcaption}
\usepackage{makecell}
\usepackage{adjustbox}

\usepackage{pifont}
\usepackage{xspace}

\definecolor{Gray}{gray}{0.9}
\definecolor{lgray}{gray}{0.95}
\definecolor{LightCyan}{rgb}{0.92,0.92,1}
\definecolor{lyellow}{rgb}{1,1,0.92}
\definecolor{lgreen}{rgb}{0.92,1,0.95}
\definecolor{llblue}{rgb}{0.92,0.93,0.95}
\definecolor{lred}{rgb}{1,0.85, 0.85}
\definecolor{tabhighlight}{HTML}{e5e5e5}

\usepackage{times}
\usepackage{latexsym}
\usepackage{amsmath}
\usepackage[T1]{fontenc}

\usepackage[utf8]{inputenc}

\usepackage{microtype}

\usepackage{inconsolata}

\usepackage{graphicx}
\usepackage{cleveref}
\usepackage{listings}

\hypersetup{
  colorlinks=true,
  urlcolor=blue,
  linkcolor=red
}

\title{LLMs Struggle with Abstract Meaning Comprehension More Than Expected}

 \author{
Hamoud Alhazmi$^{1}$ \and Jiachen Jiang$^{1}$ \\
$^{1}$Department of Computer Science and Engineering, The Ohio State University, Columbus, Ohio, USA
}

\begin{document}
\maketitle
\begin{abstract}
Understanding abstract meanings is crucial for advanced language comprehension. Despite extensive research, abstract words remain challenging due to their non-concrete, high-level semantics. SemEval-2021 Task 4 (ReCAM) evaluates models’ ability to interpret abstract concepts by presenting passages with questions and five abstract options in a cloze-style format. Key findings include: (1) Most large language models (LLMs), including GPT-4o, struggle with abstract meaning comprehension under zero-shot, one-shot, and few-shot settings, while fine-tuned models like BERT and RoBERTa perform better. (2) A proposed bi-directional attention classifier, inspired by human cognitive strategies, enhances fine-tuned models by dynamically attending to passages and options. This approach improves accuracy by 4.06\% on Task 1 and 3.41\% on Task 2, demonstrating its potential for abstract meaning comprehension. Code: \url{https://github.com/kongwanbianjinyu/semeval_2021_task4_LLM}.

\end{abstract}

\section{Introduction}
Capturing abstract meaning is a fundamental yet challenging task in natural language processing (NLP), as these non-concrete concepts are essential to tasks such as sentiment analysis, metaphor interpretation, and word sense disambiguation. Abstract words, unlike concrete terms, lack direct sensory referents (e.g., "freedom" or "justice") or belong to high-level categorical hierarchies (e.g., "animal" rather than "cat"). Despite the success of deep learning models in various NLP applications, their ability to accurately interpret abstract meanings remains limited~\cite{xu2023llms}, highlighting a gap in current language understanding systems.

The SemEval-2021 Task 4~\cite{zheng2021semeval}, known as the "Reading Comprehension of Abstract Meaning" (ReCAM), was designed to address this challenge by evaluating the extent to which machine learning models can represent and understand abstract concepts. The task presents models with passages and related questions, requiring them to select the correct answer from five abstract concept options to replace the \textit{@Placeholder} token. The ReCAM task consists of three subtasks that test the abstractness from \textit{imperceptibility},  \textit{nonspecificity} and \textit{transferability}. 
\begin{itemize}
    \item \textbf{Subtask 1} evaluates a system's ability to understand \textit{imperceptibility}, where words refer to concepts that cannot be directly perceived in the physical world (e.g., “economy” or “service” compared to more concrete terms like “tree” or “red”).
    
    \item \textbf{Subtask 2} assesses comprehension of \textit{nonspecificity}, focusing on concepts that are high in a hypernym hierarchy. For example, terms like “vertebrate” represent broader, more generalized meanings compared to specific terms like “monkey.”
    
    \item \textbf{Subtask 3} tests the model’s \textit{transferability} across types of abstractness, requiring a model trained on Subtask 1 to be evaluated on Subtask 2, and vice versa. This subtask highlights the model's ability to generalize between imperceptible and nonspecific concepts.
\end{itemize}
It has recently been suggested that large language models (LLMs), such as GPT-4~\cite{achiam2023gpt}, may exhibit "sparks of artificial general intelligence"~\cite{bubeck2023sparks}. These so-called foundation models~\cite{bommasani2021opportunities} demonstrate remarkable abilities in linguistic, factual, and commonsense reasoning. While LLMs have achieved state-of-the-art (SOTA) results across a range of tasks, an open question remains: \textit{How well do large language models perform on abstract meaning multiple-choice question-answering tasks?} 

Our research indicates that most current open-source and closed-source LLMs still face challenges in accurately comprehending abstract meanings. This observation is verified by evaluating LLMs, including Llama-3.1~\cite{touvron2023llama}, Vicuna-1.5~\cite{vicuna2023}, Qwen-2.5~\cite{qwen2023}, Gemma-2~\cite{gemma2023}, as well as closed-source models like GPT-3.5-Turbo, GPT-4o, and GPT-4o-Mini~\cite{openai2023gpt35, openai2024gpt4}, on the SemEval-2021 Task 4 Subtask 1. To adapt the multiple-choice format to generative LLM tasks, we employ a multi-choice prompting approach~\cite{robinson2023larp} that presents each question along with all answer options. The LLMs are instructed to generate a single token as the answer, chosen from options \{“0”, “1”, “2”, “3”, “4”\}. Additionally, few-shot learning is employed, providing examples to aid the models in making their selections. However, experimental results reveal that the highest accuracy achieved was only 73.60\% for Gemma-2-9B and 72.28\% for GPT-4o-Mini—both significantly trailing the benchmark's top result of 95.1\%~\cite{zheng2021semeval}.

Given the challenges in enhancing LLMs’ ability to comprehend abstract meanings, our study instead focuses on improving the abstract meaning understanding capabilities of pre-trained BERT-like language models. While advanced pre-trained language models (PLMs) such as BERT~\cite{devlin2018bert}, RoBERTa~\cite{liu2019roberta}, and DeBERTa~\cite{he2020deberta} have demonstrated outstanding performance across a range of NLP tasks, they frequently encounter difficulties in generalizing abstract meanings across varied abstraction types. For example, while these models can be trained to recognize intangible concepts like "freedom," they often struggle with broader, hierarchical abstractions, such as distinguishing between terms like "animal" and "mammal." This lack of generalization capability reveals inherent limitations in PLMs when it comes to abstract comprehension, even with fine-tuning. 

To address these limitations, our study introduces a novel bi-directional attention classifier inspired by human cognitive strategies for understanding abstract meanings. When interpreting abstract concepts, humans often engage in a two-step process: (1) they first re-examine the passage, focusing on evidence that aligns with the details provided in the question and answer options; (2) they then revisit the question and answer options, using the context from the passage to identify the correct answer while eliminating incorrect options. Our bi-directional attention classifier is designed to emulate this process. It employs self-attention to capture relationships between the question and answer choices. For the first step, we treat the passage as the query, with the question and answer options serving as the keys and values. In the second step, the question and answer options function as the query, while the passage is used as the key and value. These two attention mechanisms are then fused to create a comprehensive understanding. By allowing the model to dynamically attend to both the question and the answer options in this structured manner, our classifier enhances the ability of PLMs to grasp nuanced abstract meanings. This approach results in a notable 4.06\% accuracy improvement on Task 1 and a 3.41\% improvement on Task 2 when finetuning the pretrained model on this task. Overall, our contributions are as follows:
\begin{itemize}
    \item We reveal that most existing open-source and closed-source LLMs still struggle with abstract meaning comprehension, showing significant performance gaps in this area.
    
    \item We propose a novel bi-directional attention classifier that dynamically attends to both the passage and the question-answer options. This approach significantly improves finetuned BERT model performance, achieving a 4.06\% accuracy increase on Task 1 and a 3.41\% increase on Task 2.
    
    \item Experimental results demonstrate that combining the ELECTRA encoder with our bi-directional attention classifier achieves the best performance, ranking within the top 3 on the SemEval-2021 Task 4 benchmark.
\end{itemize}

\section{Related Works}
This section focuses on some of the systems which achieved very good performance in the SemEval-2021 Task 4~\cite{zheng2021semeval} for reading comprehension of abstract meaning. The Gated-Attention (GA) Reader, proposed by~\cite{dhingra2016gated}, leverages a multi-hop architecture with a gated attention mechanism, enabling it to build query-specific token representations for improved accuracy in reading comprehension tasks.

The TA-MAMC Attention system~\cite{zhang2021ta} achieved top performance Subtask 1 and second-top in Subtask2 and Subtask 3. The system combines ELECTRA-based models with task-adaptive pretraining and a multi-head attention classifier, achieving high accuracy in distinguishing abstract concepts. Key findings include the benefit of task-specific pretraining using news datasets, which improved performance significantly, and the use of wrong answer ensemble techniques to enhance prediction accuracy. The system’s approach highlights the advantage of adaptive pretraining and multi-head attention in handling abstract comprehension tasks, which is relevant for applications in machine reading comprehension and similar complex linguistic tasks.

This paper ~\cite{wang2021pingan} demonstrated PINGAN Omini-Sinitic system, which achieved the second-top performance in Subtask 1 and top performance in Subtask 2 and Subtask 3. The system used a pre-trained ELECTRA discriminator model, fine-tuned with innovative techniques such as upper attention and auto denoising, to effectively choose abstract words in a cloze-style format. These approaches enabled better handling of long sequences and significantly improved contextual understanding. Experimental results showed the system’s superior accuracy in tasks involving abstract concept identification, showcasing ELECTRA's potential for complex reading comprehension tasks.

\section{Methods}
In this section, we introduce the methods we used to solve Semeval2021-Task4. First, we show one instance of our tasks in \Cref{sec: problem}. In \Cref{sec:LLM}, we utilize the LLM's zero-shot or few-shot ability to directly evaluate on the validation set. In \Cref{sec:finetune}, we finetune the 
pretrained language models encoders of Robert-a and Electra on the training set and evaluate on the validation set. 

\subsection{Problem Setup}
\label{sec: problem}
Here is an instance of training dataset,
\begin{itemize}
    \item "Article": "... observers have even named it after him, ``Abenomics". It is based on three key pillars -- the "three arrows" of monetary policy, fiscal stimulus and structural reforms in order to ensure long-term sustainable growth in the world's third-largest economy. In this weekend's upper house elections, ...."
    \item "Question": "Abenomics: The @placeholder and the risks"
    \item "Option0": "chances",
    \item "Option1": "prospective",
    \item "Option2": "security",
    \item "Option3": "objectives",
    \item "Option4": "threats",
    \item "Label": 3
\end{itemize}
where "Article" provides the context for the question; "Question" models are required to answer; "Options" are five answer options for the question. Model are required to select the true answer from 5 options. "Label" is index of the answer in options.

\subsection{Leveraging Large Language Models for Multiple Choice Question Answering}
\label{sec:LLM}
In this section, we examine the ability of existing large language models (LLMs) to handle multiple-choice question answering and discuss the necessity of prompting for this task. LLMs are primarily designed for \textit{generative tasks}, where they generate answers based on a passage and question, rather than selecting from predefined options. In contrast, multiple-choice question answering is a \textit{selective task}, in which the model must choose the best answer from a set of candidates. To adapt LLMs for selective tasks, we first convert these tasks into a generative format by constructing a prompt template using samples from the dataset. To identify the optimal prompting approach for this task, we explore various techniques in \Cref{subsubsec:prompting_style} and find that a generative prompting format provides the best results. Additionally, to further enhance performance, we implement few-shot learning, as discussed in \Cref{subsubsec:few-shot}.

\subsubsection{Prompting Style}
\label{subsubsec:prompting_style}
There are three kinds of prompting styles for multi-choice question answering: Fill Back Echo Prompting, Complete Echo Prompting and Multi Choice Prompting.

\textbf{Fill Back Echo Prompting} 
A question is presented to a large language model (LLM), and each candidate answer is scored independently by the model. The selected answer is the one assigned the highest probability. In this process, the LLM does not view all answer choices simultaneously. Instead, each choice is inserted back into the original text at a specified \texttt{@Placeholder} location. The model then echoes the prompt, allowing us to obtain the predicted log-probabilities as it generates each word. Specifically, we use the log-probability associated with the \texttt{@Placeholder} location. This echoing process is repeated for each answer option, and ultimately, the option with the highest probability is selected as the final answer.

\textbf{Complete Echo Prompting} Following the same process as before, the large language model (LLM) does not view all answer choices at once. The key difference in this approach is that, instead of placing each answer choice within the original text at a \texttt{@Placeholder} location, we append each answer option to the end of the text. The LLM then echoes the prompt, producing predicted log-probabilities for each word in the sequence. Here, we focus on the log-probability at the final token position. Since placing answers at the end may lead to biases—particularly due to common or uncommon tokens or sequences of varying length—we normalize the probabilities for each option. The final answer selected is the one with the highest normalized probability.

\textbf{Multi Choice Prompting} First, the system prompt is added at the beginning:\textit{"Given the article below and the corresponding question, you are expected to choose the correct answer from five candidates to fill the \texttt{@placeholder} in cloze-style machine reading comprehension tasks. Output the answer as a single number, choosing an option from [0,1,2,3,4] that best fits the \texttt{@placeholder} in the question. I will provide you with a few-shot examples to help you understand the task better."} In this setup, the LLM is provided with all answer choices at once. The task is structured for the model to generate only a single word corresponding to one of the options: [``0'', ``1'', ``2'', ``3'', ``4'']. The selected answer is the option assigned the highest probability.

Overall, GPT-3 \cite{brown2020language} proposes a method for multiple-choice question answering using cloze prompting, which resembles the initial two prompting styles: fill-back-echo prompting and complete-echo prompting. However, as noted by \cite{robinson2023leveraginglargelanguagemodels}, cloze prompting lacks robustness; the probabilities assigned to answers can be skewed, particularly by common or uncommon tokens or sequences of differing lengths. Consequently, multi-choice prompting emerges as a more reliable approach than the earlier methods. Therefore, all subsequent experiments are conducted using multi-choice prompting.

\subsubsection{Few-Shot Learning}
\label{subsubsec:few-shot}
Few-shot learning allows large language models (LLMs) to perform better on multiple-choice question answering by providing a few labeled examples, which guide the model’s understanding of task-specific patterns. Unlike zero-shot learning, where the model relies solely on pre-trained knowledge, few-shot learning offers context about the question structure and correct answers. This approach helps the model better interpret subtle differences between choices, calibrate probability scores for each option, and improve accuracy by learning from patterns in examples. Overall, we conduct experiments using zero-shot, one-shot, and two-shot settings, observing a consistent increase in performance as more examples are provided.

\subsection{Finetuning on Pretrained Encoder-like Models}
\label{sec:finetune}
Deep learning NLP models require extensive data, but task-specific datasets with only thousands of labeled examples are often insufficient. To address this, researchers have developed pre-training techniques using vast amounts of unannotated web text. These general-purpose pre-trained models, like ReCAM, can then be fine-tuned on smaller task-specific datasets to tackle tasks such as Abstract Meaning Reading Comprehension. For this part, we finetune the pretrained language encoders
RoBERT-a \cite{liu2019roberta} and ELECTRA \cite{clark2020electra} on the training set as the baseline results. To further improve the finetuning performance, we introduce the Bi-Directional Attention classifier as shown in \Cref{bi-directional}.

\subsubsection{Bi-Directional Attention Classifier Design}
\label{bi-directional}
In this part, we would add an extra attention layer which is Bi-Directional Classifiers module as described in  \cite{DBLP:journals/corr/NguyenRSGTMD16}. Basically, it involves 1) splitting the output sequence from the encoder into question-answer sequence and passage sequence; 2) calculating two attention representations from the two sequences, one from the passage attending the question-answer, the other vice versa; 3) concatenate the two attention representations together after individually mean-pooled; 4) The representations would be sent to the classifier. The answer option with the highest probability is picked as the predicted answer. We would use the Cross Entropy function between the ground truth and the predicted probabilities to compute the loss.

\begin{figure}[!ht]
\includegraphics[width=0.5\textwidth]{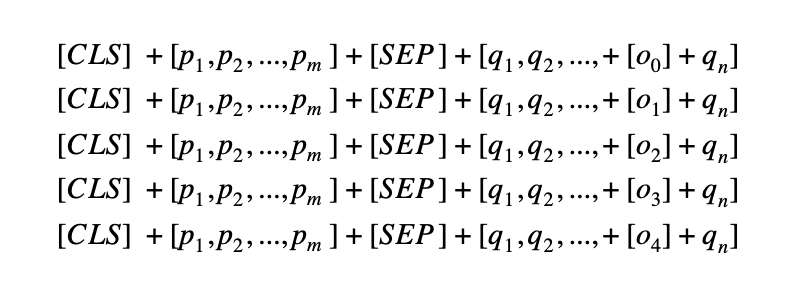}
\centering
\caption{The concatenation of passage, question and five options.}
\label{concanate}
\end{figure}

\textbf{Encoder} The encoder would encode input tokens into representations. The encoder can either be context-free or context-based. Context-free models like Word2Vec would generate a single word embedding representation (a vector of numbers) for each word in the vocabulary. The same word may have different meanings in different context. The context-based models like Pretrained LMs can generate a representation of each word that is based on the other words in the sentence, which gets a better understanding of words in the context. Therefore we choose Pretrained LMs, e.g. ROBERTa and ELECTRA, as our encoder. We fill the \textbf{@Placeholder} in the question with options and concatenate it with the  corresponding passage form one sequence and then feed it into the encoder. Let $P = [p_1,p_2,...,p_m]$, $Q = [q_1,q_2,...,q_n]$, $O = [o_0,o_1,o_2,o_3,o_4]$ be passage, question and options, where $p_i$,$q_i$ and $o_i$ are token ids in passage, question and options. The concatenation of $P,Q,O$ is shown in \Cref{concanate}.
The passage would be truncated or padding to ensure the input length of the token ids would be $5 \times 256$. The encoding output $E$ has a form $[e_1,...,e_{5 \times 256}]$, where $e_i$ is a vector of fixed dimension $d_{hidden\_size}$ that represents
the respective token.

\begin{figure}[!ht]

\centering
\includegraphics[width=0.5\textwidth]{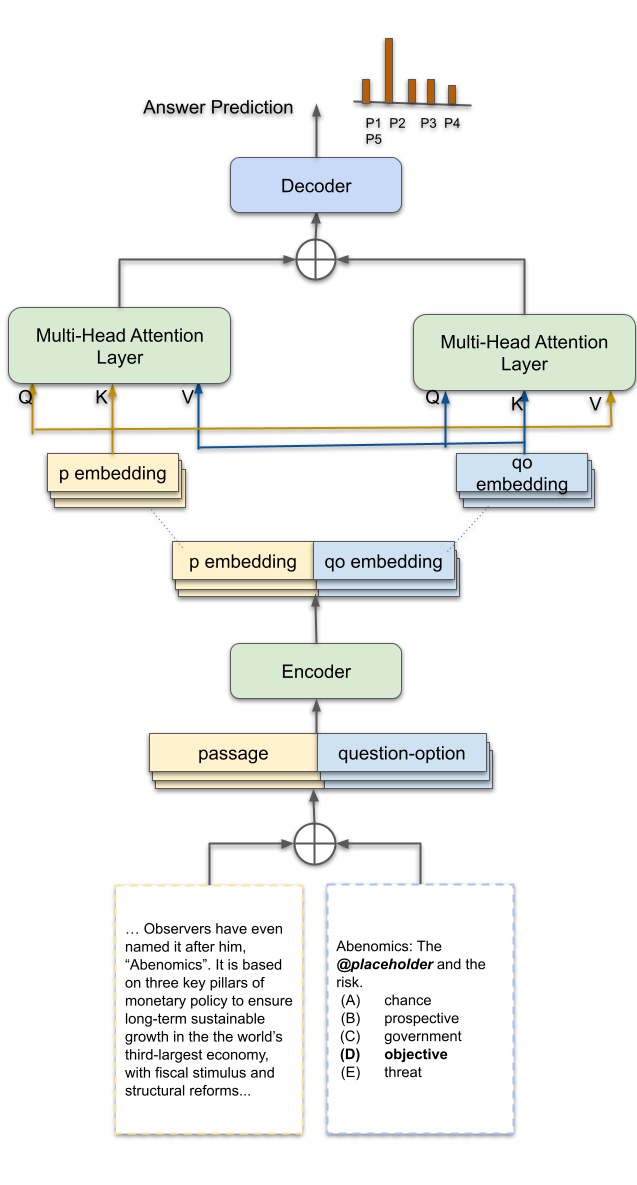}
\caption{An overview of the overall architecture of Bi-Directional Attention}
\end{figure}

\textbf{Bi-Directional Attention} The attention would use two multi-head attention layers in parallel and calculate the attention representations in a bi-directional way. Firstly, it would separate the representation from the encoder into passage embedding as $E_P$ and question-option embedding as $E_{QO}$. Then for one multi-head attention layer, it would take $E_P$ as $Query$ and $Key$ and take $E_{QO}$ as $Value$. For the other one, it would take $E_{QO}$ as $Query$ and $Key$ and take $E_{P}$ as $Value$.
\begin{equation}
\begin{aligned}
    \textit{MHA}_1 &= \textit{Attn}(E_P,E_P,E_{QO}) \\
    \textit{MHA}_2 &= \textit{Attn}(E_{QO},E_{QO},E_P) 
\end{aligned}
\end{equation}
\begin{equation}
\begin{aligned}
    \textit{Bi-Attn}(E_P,E_{QO}) &= \textit{POOL}(\textit{MHA}_1,\textit{MHA}_2)
    \end{aligned}
\end{equation}
Where $\textit{MHA}_2$ and $\textit{MHA}_2$ are output of the two multi-head attention layers. $\textit{POOL}()$ uses mean pooling to pool the sequence outputs, and $\textit{Bi-Attn}()$ is our Bi-Directional Attention module.

\textbf{Decoder} Our model decoder takes the outputs of Bi-Directional Attention and computes
the probability distribution over answer options.
\begin{equation}
    O = \textit{Bi-Attn}(E_P,E_{QO}) \\
\end{equation}
\begin{equation}
\begin{aligned}
    P(O_{correct}|P,QO) = \textit{Softmax}(\textit{DropOut}(W^TO))
\end{aligned}
\end{equation}
Where $O$ are output of Bi-Directional Attention layer, and $W^T$ is a learnable parameter in linear layer. $DropOut()$ layer adds 5 DropOut layer with dropout rate of 0.5 to prevent overfitting. $P(O_{correct}|P,QO)$ is the probability of the 5 options. Highest probability indicates the correct answer.

\subsubsection{Finetuning}

We implement our method in three main steps:

\begin{itemize}
    \item \textbf{Data Pre-processing}: For each passage, we pair it with a summary containing five candidate answers. Each candidate is substituted into the summary to create multiple complete sentences. These option-filled sentences are then concatenated with the passage tokens as input, enclosed by the [CLS] token at the start and [SEP] tokens for separation.
    
    \item \textbf{Task-adaptive Pretraining}: To improve embeddings, we apply task-adaptive pretraining on language models (LMs), as recommended by \citet{gururangan2020don}. While most LMs are trained on general corpora like Wikipedia, task-adaptive pretraining fine-tunes the model using domain-specific data (e.g., CNN/Daily Mail) to align it more closely with the target domain.

    \item \textbf{Fine-tuning}: We fine-tune our pre-trained language model (PLM) encoder on the ReCAM dataset. Using PyTorch, we load pre-trained language models and apply the AdamW optimizer for fine-tuning. We select the optimal learning rate based on the development set and set a small batch size to accommodate the GPU memory limits of Google Colab.
\end{itemize}

\section{Experiments}

In this section, we present a series of experiments to evaluate the performance of various large language models (LLMs) and fine-tuned BERT-based models on multiple-choice question answering tasks. We begin by assessing zero-shot and few-shot capabilities of LLMs and then compare these results with fine-tuned BERT variants on three tasks.

\subsection{LLM Results}

We first evaluate the zero-shot performance of several LLMs using a generative prompting approach. The results are summarized in \Cref{tab:zero_shot_accuracy}, where we report the zero-shot accuracy for each model. GPT-4o-Mini achieves the highest zero-shot accuracy (65.83\%), followed closely by Gemma 2 -9B and GPT4-o. Among the tested models, GPT-3.5 Turbo and Vicuna 1.5 7B show lower performance, indicating substantial variance in zero-shot capabilities across different LLMs.

\begin{table}[h!]
\centering
\caption{Zero-shot accuracy of different LLMs.}
\label{tab:zero_shot_accuracy}
\begin{tabular}{|l|c|}
\hline
\textbf{Model} & \textbf{Zero-shot Accuracy} \\ \hline
GPT-4o-Mini & \textbf{65.83\%} \\ \hline
Gemma-2-9B & 64.76\% \\ \hline
GPT4-o & 64.40\% \\ \hline
Qwen2.5-7B & 59.02\% \\ \hline
Meta-Llama-3.1-8B & 52.45\% \\ \hline
Vicuna 1.5 7B & 24.49\% \\ \hline
GPT3.5 Turbo & 20.91\% \\ \hline
\end{tabular}
\end{table}

Next, we further evaluate the top-performing open-source model, Gemma-2-9B, and the top-performing closed-source model, GPT-4o-Mini, under zero-shot, one-shot, and two-shot settings to assess how additional examples impact performance. As shown in \Cref{tab:few_shot_results}, both models demonstrate improved accuracy with more examples, with Gemma-2-9B achieving 73.60\% accuracy in the two-shot setting, the highest overall.

\begin{table}[h!]
\centering
\caption{Few-shot learning results for top-performing models.}
\label{tab:few_shot_results}
\resizebox{\linewidth}{!}{
\begin{tabular}{|l|c|c|c|}
\hline
\textbf{Model} & \textbf{Zero-Shot} & \textbf{One-Shot} & \textbf{Two-Shot} \\ \hline
Gemma 2 -9B & 64.76\% & 70.25\% & \textbf{73.60\%} \\ \hline
GPT-4o-Mini & 65.83\% & 71.45\% & 72.28\% \\ \hline
\end{tabular}
}
\end{table}

\subsection{Fine-tuned BERT Results}

In addition to LLM evaluations, we assess the performance of fine-tuned BERT-based models on three different tasks to understand the advantages of task-specific pretraining. Then we compare the performance of adding the Uni-Directional Attention and Bi-Directional Attention module.

\begin{table}[h!]
\centering
\caption{Accuracy of fine-tuned BERT-based models on three tasks.}
\resizebox{\linewidth}{!}{%
\begin{tabular}{|l|c|c|c|}
\hline
\textbf{Pretrained Model} & \textbf{Task1} & \textbf{Task2} & \textbf{Task3} \\ \hline
RoBERTa-large & 64.47\% & 70.47\% & 68.47\% \\ \hline
ELECTRA-large & 85.89\% & 88.00\% & \textbf{89.06\%} \\ \hline
\end{tabular}%
}

\label{tab:finetuned_bert_results}
\end{table}

\Cref{tab:finetuned_bert_results} shows the results for RoBERTa-large and ELECTRA-large, with ELECTRA-large achieving notably higher accuracy across all tasks, peaking at 89.06\% for Task 3. These findings highlight the benefits of fine-tuning pre-trained models on specific datasets.

And then we implement Uni-Directional Attn and Bi-Directional Attn into the structure and do several experiments on each of the  tasks and finally take an average accuracy to avoid unfairness.
Table \ref{tab:mamc_duma_acc_1}, \ref{tab:mamc_duma_acc_2}, \ref{tab:mamc_duma_acc_3} shows the accuracy after applying the Uni-Directional Attn layer and Bi-Directional Attn layer. Both methods have a enhancement compared to only ELECTRA-large model in 3 tasks. On average over all tasks, the accuracy improves by 0.86\% on Uni-Attn and 3.00\% on Bi-Attn.  \\
\begin{table}[!ht]
\centering
\caption{Validation Accuracy For task1 using Uni-Attn and Bi-Attn.}
\resizebox{\linewidth}{!}{
\begin{tabular}{ |c|c|c|c| } 
\hline
\textbf{Model} & \textbf{Task1} \\
\hline
ELECTRA-large & 85.89\%  \\
\hline
ELECTRA-large + Uni-Attn & 87.08\% (+1.19\%) \\
ELECTRA-large + Bi-Attn & 89.95\% (+4.06\%) \\
\hline
\end{tabular}
}

\label{tab:mamc_duma_acc_1}
\end{table}\\

\begin{table}[!ht]
\centering
\caption{Validation Accuracy For task2 using Uni-Attn and Bi-Attn.}
\resizebox{\linewidth}{!}{
\begin{tabular}{ |c|c|c|c| } 
\hline
\textbf{Model} & \textbf{Task2} \\
\hline

ELECTRA-large & 88.00\%  \\
\hline
ELECTRA-large + Uni-Attn & 89.29\% (+1.29\%) \\
ELECTRA-large + Bi-Attn & 91.41\% (+3.41\%) \\
\hline
\end{tabular}
}

\label{tab:mamc_duma_acc_2}
\end{table}

\begin{table}[!ht]
\centering
\caption{Validation Accuracy For task3 using Uni-Attn and Bi-Attn.}
\resizebox{\linewidth}{!}{
\begin{tabular}{ |c|c|c|c| } 
\hline
\textbf{Model} & \textbf{Task3} \\
\hline

ELECTRA-large & 89.06\%  \\
\hline
ELECTRA-large + Uni-Attn & 89.18\% (+0.12\%) \\
ELECTRA-large + Bi-Attn & 90.59\% (+1.53\%) \\
\hline
\end{tabular}
}

\label{tab:mamc_duma_acc_3}
\end{table}

\subsection{Comparison of Approaches}
\begin{figure}[!ht]

\centering
\includegraphics[width=0.5\textwidth]{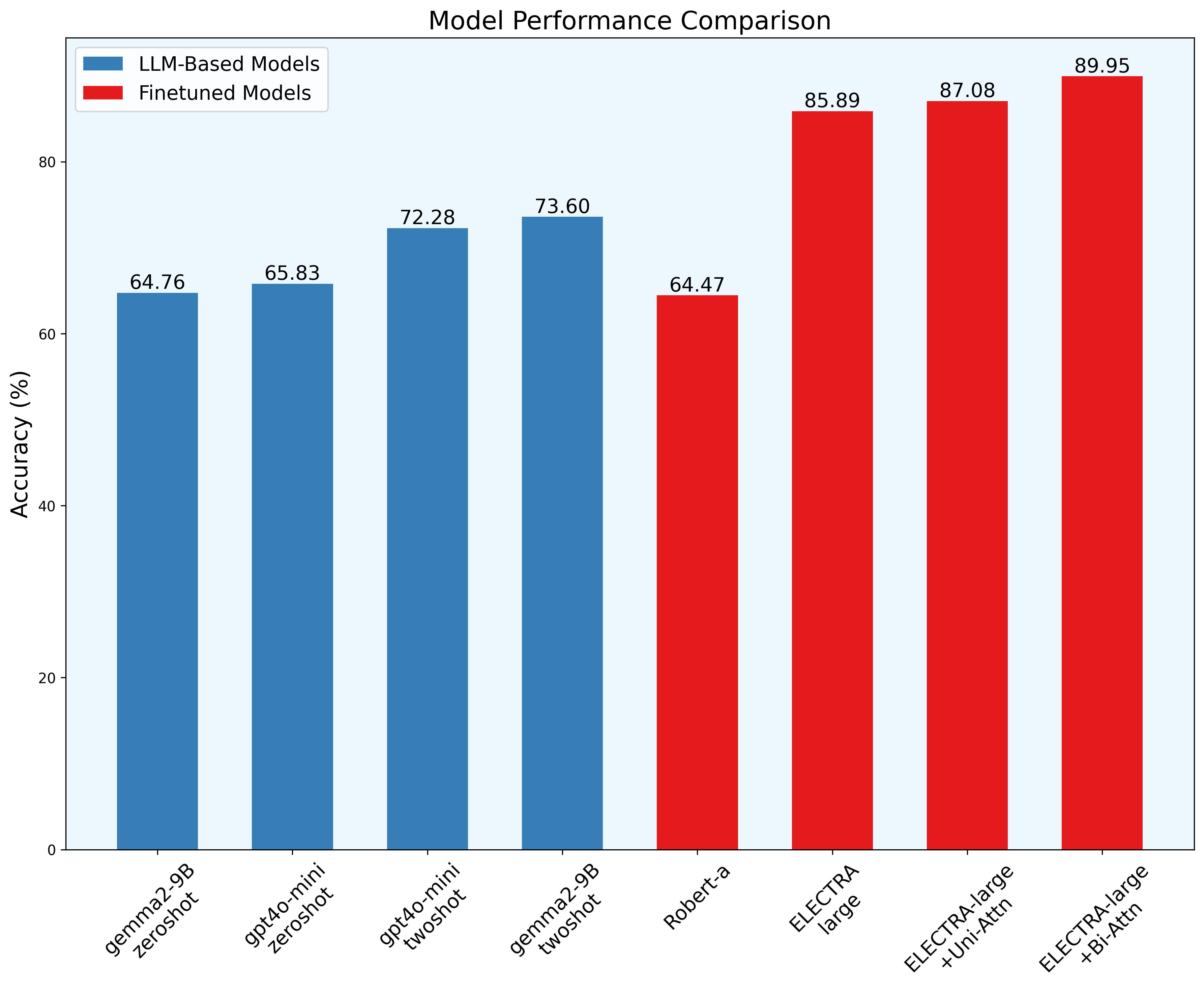}
\caption{Comparison of LLM-based model v.s. finetuned models on tasks1. The ELELCTRA with Bi-Directional Attention acheieves the best performance over all models. }
\label{fig:comparison}
\end{figure}

We compare the LLM-based models and fine-tuned models as shown in \Cref{fig:comparison}. We can draw two conclusions: First, the fine-tuned models outperform LLM-based models on average. Second, Bi-Directional Attention brings more performance gains than both the baseline and Uni-Directional Attention.

\subsection{Future Works}
In addition, when focusing on the result of task 3, we could notice that the high accuracy is due to the contribution of ELECTRA rather than Uni-Directional Attention or Bi-Directional Attention. In comparison with task 1 (+4.06\%) and task 2 (+3.41\%), we get lower enhancement of accuracy in task 3 (+1.53\%). In order to tackle with this problem, we mainly proposed 3 procedures below for better generalization:
\begin{itemize}
    \item \textbf{Data repartition} Data repartitioning (mix the train/dev sets, and randomly split into new train/dev sets by 8:2 or 9:1) aims to smooth the distribution difference among different train/dev data partition.
    \item \textbf{Data Augmentation} Augmenting the task data itself for fine-tuning, to mask different word than the original gold option (if there exists) using Task-adaptive Pretraining. The accurracy remains almost the same after adding the task augmented data. This suggests that our automatic augmentation method makes lower quality samples than the labelling data, while not too noisy that it can contribute to the robustness of the model.
    \item \textbf{Weight Averaging} Stochastic Weight Averaging \cite{izmailov2019averaging} can be done across multiple checkpoints in the same run to get better generalization as well, so we also add this method to list.
    
    \item \textbf{Negative Augmentation}  According to \citealp{DBLP:journals/corr/abs-2002-05709}, stronger negative samples will help the model learning with better performance. So we plan to generate some negative words using the pre-trained LMs to help train the models. Specifically, we replace the \textit{@placeholder} with [MASK] to reconstruct the input and ask the BERT model  to predict the word token at the [MASK]. These generated words are used as negative candidates.
\end{itemize}

\section{Conclusion}
Our system takes the large pre-trained LM ELECTRA with a Uni-Directional Attention or Bi-Directional Attention classifier on top. Firstly, we apply task-adaptive pretraining on different LMs (BERT, ROBERTa and ELECTRA) using CNN daily dataset to make the model fit the distribution in this domain. Secondly, We fine-tune them and compared the benchmark performance of different pre-trained LMs on the SemEval-2021 task 4, the result shows that ELECTRA outperforms other LMs in understanding abstractness of both the imperceptibility and nonspecificity. Therefore, we choose ELECTRA as our encoder and get a significant improvement on validation accuracy for about 20\% . Thirdly, we try two kinds of on top classifiers, both Uni-Directional Attention and Bi-Directional Attention, to replace the original softmax classifier. Uni-Directional Attention would get an improvement on validation accuracy for about 1.24\%. Bi-Directional Attention would get an improvement on validation accuracy for about 3.74\%. Finally, we evaluate the generalization ability in task3. we find that Electra model have already performed well in generalization since it can captured global contextual
sentence meaning. Our on top classifier improve the system’s performance not so much than task1 and task2.

\newpage

\appendix

\section{Pretrained Encoders}

The prominent pretrained encoders is BERT (Bidirectional Encoder Representations from Transformers), which uses a Masked Language Model (MLM) approach. By masking random words in sentences and predicting them, BERT learns bidirectional context, unlike previous models (e.g., GRU, LSTM) that process sequences in a single direction. This bidirectional training allows BERT to predict masked words based on both preceding and following words, resulting in state-of-the-art performance across diverse NLP tasks. 

\subsection{RoBERT-a}

Robustly optimized BERT approach (RoBERTa) \cite{liu2019roberta} builds on BERT, which utilizes the transformer architecture and is trained using two main objectives: masked language modeling (MLM) and next sentence prediction (NSP).

\textbf{Masked Language Model (MLM)}: In BERT, 15\% of the tokens in each input sequence are randomly replaced with the special token [MASK]. The model then predicts these masked words based on surrounding context, using a classification layer and a softmax function over the vocabulary.

\textbf{Next Sentence Prediction (NSP)}: NSP is a binary classification task that predicts whether two text segments are consecutive. Positive examples come from adjacent segments, while negative examples are randomly paired from different documents, making NSP beneficial for tasks involving sentence relationships.

RoBERTa \cite{liu2019roberta} enhances BERT by training longer with larger batches and more data, removing the NSP objective, training on longer sequences, and dynamically varying the masking patterns, resulting in performance that matches or exceeds that of post-BERT models.

\subsection{ELECTRA}

Several models, such as RoBERTa and XLNet, have extended BERT’s success by leveraging larger networks and datasets. However, the Efficiently Learning an Encoder that Classifies Token Replacements Accurately (ELECTRA) \cite{clark2020electra} model introduces a new training approach that achieves comparable or better performance with significantly less computational cost. 

Instead of using Masked Language Modeling (MLM), ELECTRA employs a "replaced token detection" technique, which includes both a generator and a discriminator. The generator, an MLM model, predicts the value of masked tokens, while the discriminator identifies which tokens in the sequence are original and which are replaced by the generator. Once training is complete, only the discriminator is retained as the ELECTRA model, which can then be used as an encoder for downstream tasks. This approach is more efficient than MLM, as it requires the model to evaluate every token in a sample rather than focusing solely on [MASK] tokens. Importantly, ELECTRA’s architecture is not a GAN, as the generator does not optimize to mislead the discriminator.

\section{Attention}
An attention function can be described as mapping a query and a set of key-value pairs to an output, where the query, keys, values, and output are all vectors. The output is computed as a weighted sum of the values, where the weight assigned to each value is computed by a compatibility function of the query with the corresponding key.
\subsection{Scaled Dot-Product Attention}
We call our specified attention "Scaled Dot-Product Attention". The input consists of queries and keys of dimension $d_k$, and values of dimension $d_v$. We compute the dot products of the query with all keys, then divide each by $\sqrt{d_k}$. After that we apply a softmax function to obtain the weights on the values.
For a set of queries, we compute the attention function simultaneously and pack results together
into a matrix Q. The keys and values are also packed together into matrices K and V . So that the output can be calculated as below:
\begin{equation}
\begin{aligned}
    Attention(Q, K, V) = softmax(\frac{QK^T}{\sqrt{d_k}})V ,
\end{aligned}
\end{equation}
where $\frac{1}{d_k}$ is the scaling factor.\\
In comparison with additive attention \cite{bahdanau2016neural_additive_attention}, dot-product attention is faster and more space-efficient in practice though they  are similar in complexity because dot-product attention can be implemented using highly optimized matrix multiplication code.

\subsection{Multi-Head Attention}
Instead of performing a single attention function with $d_{model}$-dimensional keys, values and queries,
we found it beneficial to linearly project the queries, keys and values h times with different, learned linear projections to $d_k$, $d_k$ and $d_v$ dimensions respectively. On each of these projected triples we then perform the attention function, yielding $d_v$-dimensional output values. These are concatenated and projected again and output the final values.

Multi-head attention allows the model to jointly obtain information from different representation
subspaces, while with a single attention head averaging inhibits this. And the output can be calculated as:
\begin{equation}
\begin{aligned}
\mathrm{MultiHead}(Q, K, V) = \mathrm{concat}(head_1, head_2, \dots, head_h)W^O
\end{aligned}
\end{equation}

\begin{equation}
\begin{aligned}
head_i = \mathrm{Attention}(QW_i^Q, KW_i^K, VW_i^V)
\end{aligned}
\end{equation}

where $W_i^Q \in \mathbb{R}^{d_{\mathrm{model}} \times d_k}$, 
$W_i^K \in \mathbb{R}^{d_{\mathrm{model}} \times d_k}$, 
$W_i^V \in \mathbb{R}^{d_{\mathrm{model}} \times d_v}$, 
and $W_i^O \in \mathbb{R}^{h d_v \times d_{\mathrm{model}}}$.

\section{Uni-Directional Attention}

In this part, we introduce the ablation of uni-directional attention as described in \cite{vaswani2017attention}. While putting into usage in MRC, we firstly take the encoder output as input Q, K, V of muti-head attention layer. Then calculate the attention representations from the concatenated embeddings. Finally take the out put as the input of decoder and get the option with highest possible rate as our prediction. And the structure is showed in Figure \ref{Uni-Directional Attention} below.
\begin{figure}[!ht]
\includegraphics[width=0.4\textwidth]{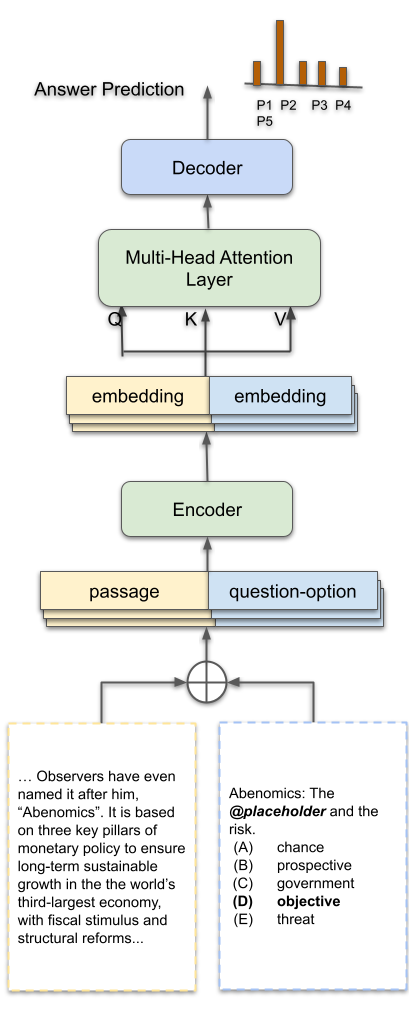}
\centering
\caption{An overview of the overall architecture of Uni-Directional Attention}
\label{Uni-Directional Attention}
\end{figure}

\section{Details of Experiment Setting}

\subsection{Dataset}

\begin{itemize}
    \item \textbf{ReCAM}. Dataset for the SemEval-2021 Task 4. Data is stored one-question-per-line in json format, including article, question, options and label. In Subtask 1, the training/ trail/ development/ test contains 3,227/1,000/837/2,025 instances. In Subtask 2, the training/ trail/ development/ test contains 3,318/1,000/851/2,017 instances.

    \item \textbf{CNN/Daily Mail}. It consists 300k unique news articles as written by journalists at CNN and the Daily Mail. We would use it to implement task-adaptive pretraining.
\end{itemize}

\subsection{Network Layers}
Table \ref{tab:electra_network_layers} shows our Electra network structure, the parameter amount and whether they are trainable. We use Electra model as encoder and a linear trainable layer as classifier.
\begin{table}[!ht] 
\centering
\caption{Network layers of Electra and parameter size }
\resizebox{\linewidth}{!}{
\begin{tabular}{ |c|c | c |c|} 
\hline
\textbf{Name} & \textbf{Type} & \textbf{Params} &\textbf{Trainable}\\
\hline
 Electra & ElectraModel &  334 M & No\\
Classifier & Linear &  1.0 K & Yes\\
\hline
\end{tabular}

}
\label{tab:electra_network_layers}
\end{table}

Table \ref{tab:mamc_network_layers} shows network layers of Electra + Uni-Directional Attention model. In this structure, a trainable Muti-head attention layer is added after encoder layer, the parameter added is only 4.2M. And a dropout layer is added to avoid overfitting as well. 
\begin{table}[!ht]
\centering
\caption{Network layers of Electra + Uni-Directional Attention and parameter size.}
\resizebox{\linewidth}{!}{
\begin{tabular}{ |c|c | c |c|} 
\hline
\textbf{Name} & \textbf{Type} & \textbf{Params} &\textbf{Trainable}\\
\hline
 ELECTRA & ElectraModel &  334 M & No\\
 Uni-Attn & MultiheadAtt- &  4.2 M & Yes\\
      & entionLayer &  & \\
Dropouts & ModuleList &  0 M & None\\
Classifier & Linear &  1.0 K & Yes\\
\hline
\end{tabular}
}

\label{tab:mamc_network_layers}
\end{table}

Table \ref{tab:duma_network_layers} shows network layers of Electra + Bi-Directional Attention model. In this structure, a trainable Dual Multi-head Co-Attention layer is added after encoder layer. Since the co-attention layer shoulders double workload compared with that in Uni-Directional Attention, the parameter added is 8.4M. And a dropout layer is added to avoid overfitting as well.
\begin{table}[!ht]
\centering
\caption{Network layers of Electra + Bi-Directional Attention and parameter size.}
\resizebox{\linewidth}{!}{
\begin{tabular}{ |c|c|c|c| } 
\hline
\textbf{Name} & \textbf{Type} & \textbf{Params} &\textbf{Trainable}\\
\hline
 ELECTRA & ElectraModel &  334 M & No\\
 Bi-Attn & Bi-Directional Attention Layer &  8.4 M & Yes\\
Dropouts & ModuleList &  0 M & None\\
Classifier & Linear &  1.0 K & Yes\\
\hline
\end{tabular}
}

\label{tab:duma_network_layers}
\end{table}

\subsection{Hyperparameter Setting}

\begin{table}[!ht]
\centering
\caption{Hyperparameter Setting}
\resizebox{\linewidth}{!}{
\begin{tabular}{ |c|c|c| } 
\hline
\textbf{Layer} & \textbf{Hyperparameter} & \textbf{Value} \\
\hline
 &token max length & 256 \\
Tokenizer &truncation &  "only first"\\
& padding &'max length' \\
\hline
&learning rate & 1e-4  \\
&train batch size & 2 \\
&eval batch size & 2 \\
Trainer&train epochs & 1.0 \\
&val check interval & 0.2 \\
&dropout rate & 0.5 \\
&gradient accumu- & 32 \\
&lation steps & \\
\hline
 & type  & AdamW \\
Optimizer & lr & 1e-4\\
& weight decay & 0.01 \\
\hline
\end{tabular}
}

\label{tab:hyperparameter}
\end{table}

Table \ref{tab:hyperparameter} shows our fine-tuned hyperparameters. In the initial tokenizer layer, we set the token max length to 256 to limit the computation load. And we only need the truncation at the first time, so the truncation is set to "only first" option, and the padding is set to "max length" option to normalize sentence sizes.
While training, we set learning rate to 1e-4 after many tries and set batch size to 2 due to GPU limitaion. To accelerate the speed as well as the performance, we set gradient accumulation steps to 32 so that we can break GPU memory boundaries even with large batch sizes.

Finally we choose AdamW as our optimizer and set the leraning rate and weight decay seperately to 1e-4 and 0.01.

\bibliography{custom}

\end{document}